\documentclass[sigconf]{acmart}

\usepackage{tcolorbox}
\usepackage{latexsym}

\usepackage[utf8]{inputenc} 
\usepackage[T1]{fontenc}    
\usepackage{hyperref}       
\usepackage{url}
\usepackage{multirow}       
\newcommand{\mpara}[1]{\medskip\noindent{\bf #1}}

\usepackage{booktabs}       
\usepackage{amsfonts}       
\usepackage{nicefrac}       
\usepackage{tabularx}

\usepackage{makecell}
\usepackage{microtype}      
\usepackage{xcolor}         
\usepackage{tabularx}
\usepackage{graphicx}
\graphicspath{ {imgs/} }
\usepackage{svg}
\usepackage{enumitem}
\usepackage{subcaption}
\usepackage{amsthm, amsmath,algorithm2e}
\usepackage{pifont}
\usepackage{tabularx}
\usepackage{makecell}
\usepackage{xcolor}         
\usepackage{svg}
\usepackage{supertabular}
\usepackage{colortbl}
\usepackage{natbib}
\usepackage{setspace}
\newtcolorbox{fullpagebox}{
  width=\textwidth,
  height=\textheight,
  left=0pt,
  right=0pt,
  top=0pt,
  bottom=0pt,
  colback=white, 
  colframe=black, 
  boxrule=0.5pt,
  arc=0pt, 
  boxsep=5pt, 
  breakable 
}
\definecolor{blue}{RGB}{17,220,247}
\definecolor{purple}{RGB}{163,115,250}
\definecolor{caribbeangreen}{rgb}{0.0, 0.8, 0.6}

\definecolor{GREEN}{RGB}{84,130,53}
\newcommand{\cmark}{\ding{51}}%
\newcommand{\xmark}{\ding{55}}%
\newcommand{\colorg}{\cellcolor{gray!15}}

\AtBeginDocument{%
  \providecommand\BibTeX{{%
    \normalfont B\kern-0.5em{\scshape i\kern-0.25em b}\kern-0.8em\TeX}}}

\newcommand{\finqaroberta}{\textsc{FinQA-Roberta-Large}}

\newcommand{\flant}{\textsc{flan-t5-xl}}
\newcommand{\numt}{\textsc{NumT5-small}}

\newcommand{\programfc}{\textsc{Program-FC}}
\newcommand{\numclaims}{\textsc{QuanTemp}}
\newcommand{\claimdecomp}{\textsc{ClaimDecomp}}
\newcommand{\bart}{\textsc{bart-large-mnli}}
\newcommand{\gptt}{\textsc{gpt-3.5-turbo}}

\title{\numclaims: A real-world open-domain benchmark for fact-checking numerical claims}
  \author{Venktesh V}
  \email{v.Viswanathan-1@tudelft.nl}
 \affiliation{
   \institution{Delft University of Technology}
   \city{Delft}
   \country{Netherlands}
 }
 
 \author{Abhijit Anand}
  \email{aanand@L3S.de}
 \affiliation{
   \institution{L3S Research Center}
   \city{Hannover}
   \country{Germany}
 }

 \author{Avishek Anand}
  \email{avishek.anand@tudelft.nl}
 \affiliation{
   \institution{Delft University of Technology}
   \city{Delft}
   \country{Netherlands}
 }

 \author{Vinay Setty}
  \email{vsetty@acm.org}
 \affiliation{
   \institution{University of Stavanger}
   \city{Stavanger}
   \country{Norway}
 }
 
\additionalaffiliation{%
   \institution{Factiverse AI}
   \city{Stavanger}
   \country{Norway}}

\copyrightyear{2024} 
\acmYear{2024} 
\setcopyright{rightsretained} 
\acmConference[SIGIR '24]{Proceedings of the 47th International ACM SIGIR Conference on Research and Development in Information Retrieval}{July 14--18, 2024}{Washington, DC, USA}
\acmBooktitle{Proceedings of the 47th International ACM SIGIR Conference on Research and Development in Information Retrieval (SIGIR '24), July 14--18, 2024, Washington, DC, USA}
\acmDOI{10.1145/3626772.3657874}
\acmISBN{979-8-4007-0431-4/24/07}

\begin{document}

\begin{abstract}
With the growth of misinformation on the web, automated fact checking has garnered immense interest for detecting growing misinformation and disinformation. Current systems have made significant advancements in handling synthetic claims sourced from Wikipedia, and noteworthy progress has been achieved in addressing real-world claims that  are verified by fact-checking organizations as well. We compile and release \textbf{\numclaims{}}, a diverse, multi-domain dataset focused exclusively on numerical claims, encompassing comparative, statistical, interval, and temporal aspects, with detailed metadata and an accompanying evidence collection. This addresses the challenge of verifying real-world numerical claims, which are complex and often lack precise information, a gap not filled by existing works that mainly focus on synthetic claims. We evaluate and quantify these gaps in existing solutions for the task of verifying numerical claims. We also evaluate claim decomposition based methods, numerical understanding based natural language inference (NLI) models and our best baselines achieves a macro-F1 of 58.32. This demonstrates that \numclaims{} serves as a challenging evaluation set for numerical claim verification.

\end{abstract}

\ccsdesc[500]{Computing methodologies~Natural language processing}

\keywords{Fact-checking, Large Language Models, Claim Decomposition}

\maketitle

\begin{table}[htb!!] 
\centering
\begin{tcolorbox}[title= Example: Claim from \numclaims{}]
\small
\medskip\noindent \paragraph{\textbf{Claim}:}\texttt{Under GOP plan, U.S. families making ~\$86k see avg tax increase of \$794.}

\medskip\noindent \paragraph{[\textbf{Evidence}]:}  If enacted, the Republican tax reform proposal  would saddle \textcolor{magenta}{only 8 million households that earn up to \$86,100} with an average tax increase of \$794 …. \textcolor{magenta}{Only a small percentage (6.5 percent) of the nearly 122 million} households in the bottom three quintiles will actually face a tax increase.

\medskip\noindent \paragraph{[\textbf{Verdict}]:}\textcolor{teal}{\textsf{False}}

\end{tcolorbox}

\vspace{-1em}
\captionof{figure}{Example claim from \numclaims{}}\label{intro:example} 
\end{table}






\vspace{-1em}
\section{Introduction}
Online misinformation, particularly during elections, poses a significant threat to democracy by inciting socio-political and economic turmoil~\cite{vosoughi2018spread}. Fact-checking websites like \texttt{Politifact.com, Snopes.com, FullFact.org}, and others play an indispensable role in curbing misinformation. However, the scalability of manual fact-checking is constrained by limited resources. This limitation has spurred remarkable advancements in neural models for automated fact-checking in recent years~\cite{guo2022survey}, driven by the proliferation of open datasets~\cite{thorne-etal-2018-fever, popat2018declare, chen2020tabfact, multifc, schlichtkrull2023averitec}.
Crucially, within the area of fact-checking, the verification of claims involving numerical quantities and temporal expressions is of utmost importance. This is essential for countering the `numeric-truth-effect'\cite{sagara2009consumer}, where the presence of numbers can lend a false aura of credibility to a statement. Numerical claims are a significant component of political discourse. For instance, our analysis of the \textsc{ClaimBuster dataset}~\cite{hassan2017toward} reveals that a substantial $36\%$ of all check-worthy claims in U.S. presidential debates involve numerical quantities. 
Most current datasets inadequately address the verification of numerical claims, as our overview in Table~\ref{tab:datasets_overview} illustrates. A notable example is the \textsc{Feverous} dataset, where only a small fraction (approximately $10\%$) of claims necessitate numerical reasoning, and these have proven especially challenging for annotators to verify~\cite{aly2021feverous}. Our experiments further reinforce this difficulty. 
We observed that models trained on a mix of numerical and non-numerical claims underperform compared to those specifically fine-tuned on numerical claims. 

Numerical claims verification poses a unique challenge, where a fact-checking system must critically analyze and reason about the numerical data presented in both the claim and its evidence. For example, in verifying the claim shown in Figure \ref{intro:example} as `False', the NLI model needs to identify that the evidence only mentions 8 million households with incomes up to \$86k facing tax increases, contradicting the claim of tax increases for all families earning \$86k.
The existing datasets can be categorized as synthetically generated from Wikipedia and knowledge bases or real-world claims collected from fact-checking websites~(Table~\ref{tab:datasets_overview}). While, works like \textsc{ClaimDecomp}~\cite{claimdecomp}, \textsc{MultiFC}~\cite{multifc} and \textsc{AveriTec}~\cite{schlichtkrull2023averitec}, collect real-world claims, they do not particularly focus on numerical claims. The only previous work proposing a dataset for fact-checking statistical claims, by \cite{vlachos_simple_numerical_1, numerical_claims_3}, uses a rule-based system to create synthetic claims from 16 simple statistical characteristics in the Freebase knowledge base about geographical regions. There has not been a dedicated large-scale real-world open-domain diverse dataset for verifying numerical claims.

\begin{table*}[hbt!]
\centering
\caption{Comparison of \numclaims\  with other fact checking datasets. $^*$Some datasets may have some numerical claims in them, but it is not their main focus. $^\dagger$By ``Unleaked Evidence'', here we refer to gold evidence being leaked from fact-checking websites. In the table WP refers to Wikipedia, WT refers to WikiTables, FCS referes to fact-check sites and  FB referes to FreeBase}
{\begin{tabular}{l|rccccr}
\hline
\textbf{Dataset} & \textbf
{\# of } & \textbf{Claims } & \textbf{Retrieved} & \textbf{Numerical}  & \textbf{$^\dagger$Unleaked } & \textbf{Evidence}
\\ 
 &  \textbf{claims}&\textbf{Source}  & \textbf{Evidence} & \textbf{Focus$^*$}  & \textbf{ Evidence$^\dagger$} & \textbf{Corpus Size}
\\ 

\midrule
\textbf{Synthetic Claims} & & & & &  &\\
 \textsc{StatProps}~\cite{vlachos_simple_numerical_1} &4,225  & KB (FB) & KB (FB)& \cmark & N/A  & N/A\\
 
\textsc{Fever}~\cite{thorne-etal-2018-fever} &185,445 & WP & WP& \xmark &   N/A  & 5,416,568 \\
\textsf{Hover}~\cite{jiang2020hover} &26,171  & WP & WP& \xmark &   N/A  & 5,486,211 \\
\textsc{TabFact}~\cite{chen2020tabfact} &92,283 & WP & WT& \xmark &   N/A  & 16,573 \\
\textsc{Feverous}~\cite{aly2021feverous} &87,026  & WP & WT,WP& \xmark  &  N/A  & 7,221,406\\
\textsc{SciTab}~`\cite{lu-etal-2023-scitab} \ & 1,225 & arXiv (CS) & arXiv 
 & \xmark & N/A  & 1,301\\
\midrule

 \textbf{Fact-checker Claims}& & & & &  &\\
 LIAR~\cite{liar} &12,836  & Politifact & \xmark& \xmark & N/A  & N/A\\
  \claimdecomp{}~\cite{claimdecomp}  & 1,250 & Politifact & \xmark & \xmark&  \xmark  & 1,250\\
\textsc{DeClarE}~\cite{popat2018declare} & 13,525 &  FCS (4) & Web& \xmark &  \cmark  & 87,643\\

\textsc{MultiFC}~\cite{multifc} & 36,534  &  FCS (26) & Web& \xmark  &  \xmark & ~349,180\\
\textsc{QABriefs}~\cite{QABriefs}& 8,784  &  FCS (50) & Web & \xmark &  \xmark  & 21,168\\

\textsc{AVeriTeC}~\cite{schlichtkrull2023averitec} & 4,568  &  FCS (50) & Web& \xmark &  \xmark & ~ 137,040 \\
\midrule
\textbf{\numclaims{}}~(OURS) &15,514 & FCS (45)& Web & \cmark & \cmark  & 423,320\\
\bottomrule
\end{tabular}}

\label{tab:datasets_overview}
\end{table*}
In this work, we collect and release a dataset of $15,514$ real-world claims with \textbf{numeric quantities and temporal expressions} from various fact-checking domains, complete with detailed metadata and an evidence corpus sourced from the web. \textit{Numeric claims are defined as statements needing verification of any explicit or implicit quantitative or temporal content}. 
The evidence collection method is crucial in fact-checking datasets. While datasets like \textsc{MultiFC} and \textsc{DeClarE} use claims as queries in search engines like Google and Bing for evidence, methods like \textsc{ClaimDecomp} depend on fact-checkers' justifications. However, this could cause `gold' evidence leakage from fact-checking sites into the training data. To avoid this, we omit results from fact-checking websites in our evidence corpus.

Moreover, using claims as queries in search engines may miss crucial but non-explicit evidence for claim verification. To overcome this, recent works have proposed generating decomposed questions to retrieve better evidence \cite{QABriefs,claimdecomp,rani2023factify5wqa,aly-vlachos-2022-natural,claim_decompose}. In our approach, we aggregate evidence using both original claims and questions from methods like \textsc{ClaimDecomp} and \textsc{ProgramFC}. This dual strategy yields a more diverse and unbiased evidence set of \textbf{423,320 snippets}, enhancing which could be used for evaluating both the retrieval and NLI steps of fact-checking systems.

Finally, we also propose a fact-checking pipeline as a baseline that integrates claim decomposition techniques for evidence retrieval, along with a range of Natural Language Inference (NLI) models, encompassing pre-trained, fine-tuned, and generative approaches, to evaluate their efficacy on our dataset. Additionally, we conduct an error analysis, classifying the numerical claims into distinct categories to better understand the challenges they present.



\subsection{Research Questions}
In addition to collecting and releasing the dataset, we answer the following research questions by proposing a simple baseline system for fact-checking.

\mpara{RQ1}: How hard is the task of fact-checking numerical claims?

\mpara{RQ2}: To what extent does claim decomposition improve the verification of numerical claims?

\mpara{RQ3}: How effectively do models pre-trained for numeric understanding perform when fine-tuned to fact-check numerical claims?

\mpara{RQ4}: How does the size of large language models impact their performance in zero-shot, few-shot, and fine-tuned scenarios for numerical claims?

\subsection{Contributions}
\begin{enumerate}[itemsep=0pt, topsep=0pt]
    \item We collect and release a large, diverse multi-domain dataset of real-world \textbf{15,514 numerical claims}, the first of its kind, along with an associated evidence corpus consisting of \textbf{423,320} snippets.

    \item We evaluate established fact-checking pipelines and claim decomposition methods, examining their \textbf{effectiveness in handling numerical claims}. Additionally, we propose improved baselines for the natural language inference (NLI) step.

    \item Our findings reveal that NLI models pre-trained for \textbf{numerical understanding outperform generic models} in fact-checking numerical claims by up to \textbf{$11.78\%$}. We also show that smaller models fine-tuned on numerical claims outperform larger models like GPT-3.5-Turbo under zero-shot and few-shot scenarios.

    \item We also \textbf{assess the quality of questions} decomposed by \textsc{ClaimDecomp} and \textsc{ProgramFC} for numerical claims, using both automated metrics and manual evaluation.
\end{enumerate}

We make our \textit{dataset and code } available here \url{https://github.com/factiverse/QuanTemp}.


\section{Related Work}

The process of automated fact-checking is typically structured as a pipeline encompassing three key stages: claim detection, evidence retrieval, and verdict prediction, the latter often involving stance detection or natural language inference (NLI) tasks \cite{guo2022survey, Botnevik:2020:SIGIR, zeng2021automated}. In this work, we introduce a dataset of numerical claims that could be used to evaluate the evidence retrieval and NLI stages of this pipeline. This section will explore relevant datasets and methodologies in this domain.


Most current fact-checking datasets focus on textual claims verification using structured or unstructured data~\cite{zeng2021automated,thorne-vlachos-2018-automated}. However, real-world data, like political debates, frequently involve claims requiring numerical understanding for evidence retrieval and verification. It has also been acknowledged by annotators of datasets such as FEVEROUS that numerical claims are hard to verify since they require reasoning and yet only 10\% of their dataset are numerical in nature \cite{aly2021feverous}.

A significant portion of the existing datasets collect claims authored by crowd-workers from passages in Wikipedia \cite{jiang2020hover, thorne-etal-2018-fever, aly2021feverous, creditassess, schuster2021vitamin,sathe-etal-2020-automated}. Additionally, there are synthetic datasets that require tabular data to verify the claims \cite{chen2020tabfact,aly2021feverous,lu-etal-2023-scitab}, but these claims and tables may not contain numerical quantities. Recent efforts by \cite{kamoi2023wice} aim to create more realistic claims from Wikipedia by identifying cited statements, but these do not reflect the typical distribution of claims verified by fact-checkers and the false claims they contain are still synthetic.


More efforts have been made to collect real-world claims in domains like politics \cite{claimdecomp,liar,political_claims_1,multi_hop_political}, science \cite{wadden-etal-2020-fact,vladika2023scientific,wright2022generating}, health \cite{kotonya2020explainable} and climate \cite{diggelmann2021climatefever} and other natural claims occurring in social media posts \cite{Mitra_Gilbert_2021,derczynski-etal-2017-semeval}. Multi-domain claim collections like MultiFC \cite{multifc} have also emerged, offering rich meta-data for real-time fact-checking. However, none of these datasets focus on numerical claims.


Among those that focus on numerical claims, \cite{numerical_claims_3,vlachos_simple_numerical_1}, the authors propose a simple distant supervision approach using freebase to verify simple statistical claims. These claims are not only synthetic but they can be answered with simple KB facts such as Freebase. Similarly, \cite{financial_facts_1} explore the extraction of formulae for checking numerical consistency in financial statements by also relying on Wikidata. Further, \cite{quantitative_claims} explore the identification of quantitative statements for fact-checking trend-based claims. None of these datasets are representative of real-world claims.

In this work, we collect and release a multi-domain dataset which is primarily composed of numerical claims and temporal expressions with fine-grained meta-data from fact-checkers and an evidence collection. To the best of our knowledge, this is the first natural numerical claims dataset.
Another set of related work is the area of temporal information retrieval that deals exclusively with time~\cite{DBLP:journals/ftir/KanhabuaBN15,singh:2018:axiv:history}.
However, these works do not focus on temporal reasoning and instead deal with indexing and processing temporal queries~\cite{holzmann:2016:WWW:tempas,anand2011temporal,anand:2012:SIGIR:indexmaintainance,anand:2010:cikm:temporalkeywordsearch}.

Early fact checking systems simply used the claim as the query to search engine~\cite{guo2022survey,popat2018declare,multifc} or employ question answering systems~\cite{saha2020question:book}. 
These approaches may not work well if the claims are not already fact-checked. In this regard, recent works have introduced claim decomposition into questions~\cite{claimdecomp,QABriefs,schlichtkrull2023averitec,programfc}. In this paper, we evaluate the effectiveness of \textsc{ClaimDecomp}~\cite{claimdecomp} and \textsc{ProgramFC}~\cite{programfc} methods for numerical claims.

We follow the established fact-checking pipeline using evidence and claims as input to NLI models to predict if the claims are supported, refuted or conflicted by the evidence~\cite{guo2022survey}. We use BM25~\cite{robertson2009probabilistic} for evidence retrieval followed by re-ranking and explore various families of NLI models.

\section{Dataset Construction}
\begin{figure*}[ht!!!]
    \centering
    \includegraphics[width=0.8\linewidth]{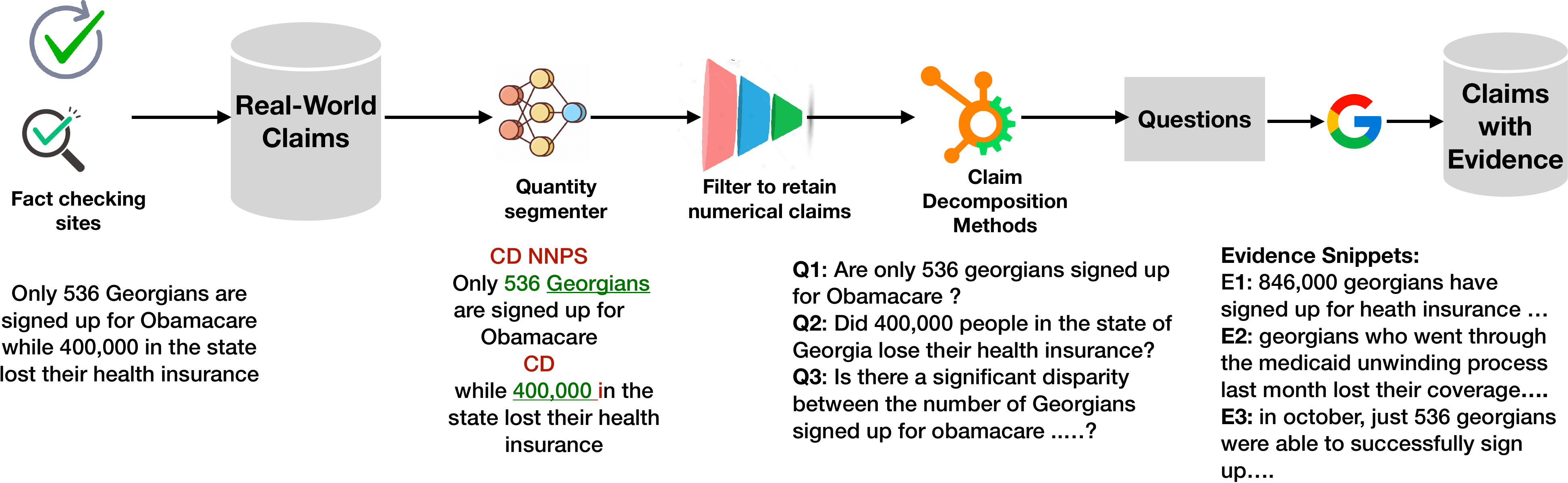}
    \caption{\numclaims{} Construction Pipeline}
    \label{fig:data_construction}
\end{figure*}

In this section, we describe an overview of the dataset collection process as shown in Figure~\ref{fig:data_construction}. 
\subsection{Collecting Real-world Claims}
\label{section:data}

We first collect real-world claims curated by professional fact-checkers at fact-checking organizations via Google Fact Check Tool APIs\footnote{\url{https://toolbox.google.com/factcheck/apis} available under the CC-BY-4.0 license.}. The full collection of fact-checks consists of 278,636 fact-checks\footnote{As of 15th Nov 2023.}. After filtering non-English fact-checks, it amounts to 45 organizations worldwide with 139,926 fact-checks. Next, we identify quantitative segments (Section~\ref{section:quantitative}) from the claims and only retain claims that satisfy these criteria, which amounts to 15,514 claims. Finally, we collect evidence for the claims (Section~\ref{section:evidence}).
\begin{figure*}
\begin{minipage}{0.32\textwidth}
\centering
\captionof{table}{Top fact-checking domains}
\label{tab:claim_sources}
\begin{tabular}{lr}
\hline
\textbf{Claim Source} & \textbf{\#Occurences}  \\

\midrule
Politifact & 3,840 \\
Snopes & 1,648 \\
AfP & 412 \\
Africacheck & 410 \\
Fullfact & 349 \\
Factly & 330 \\
Boomlive\_in & 318 \\
Logically & 276 \\
Reuters & 235 \\
Lead Stories & 223 \\

\bottomrule
\end{tabular}

\end{minipage}
\begin{minipage}{0.32\textwidth}
\centering
\captionof{table}{Top claim source countries.}
\label{tab:georgraphical_sources}
\begin{tabular}{lr}
\hline
\textbf{Country} & \textbf{\#Occurences}  \\

\midrule
USA & 6,215 \\
India & 1,356 \\
UK & 596 \\
France &  503 \\
South Africa & 410 \\
Germany & 124 \\
Philippines & 103 \\
Australia & 65 \\
Ukraine & 35 \\
Nigeria & 17 \\
\bottomrule
\end{tabular}

\end{minipage}
\begin{minipage}{0.32\textwidth}
\centering
\captionof{table}{Top evidence domains.}
\label{tab:evidence_domains}
{\begin{tabular}{lr}
\hline
\textbf{Category} & \textbf{\#Occurences}  \\

\midrule
\texttt{en.wikipedia.org} & 28,124 \\
\texttt{nytimes.com} & 8,430 \\
\texttt{ncbi.nlm.nih.gov} & 8,417 \\
\texttt{quora.com} & 4,967 \\

 
 \texttt{cdc.gov} & 3,987 \\

  
   \texttt{statista.com} & 3,106 \\
 \texttt{youtube.com} & 2,889 \\
 \texttt{who.int} & 2,557 \\
 \texttt{cnbc.com} & 2,448 \\
 \texttt{investopedia.com} & 1977 \\
\bottomrule
\end{tabular}}

\end{minipage}
\end{figure*}



One of the challenges of collecting claims from diverse sources is the labelling conventions. To simplify, we standardize the labels to one of \textit{True, False or Conflicting} by mapping them similar to \cite{schlichtkrull2023averitec}.  We also ignore those claims with unclear or no labels.

\subsection{Dataset Statistics}

\begin{table}
    \centering
    \caption{\numclaims{}~dataset distribution}
    \begin{tabular}{l|rrrr}
    \hline
        Split & True & False & Conflicting & Total \\
        \midrule
        Train & 1,824 &  5,770 & 2,341 & 9,935 \\
        Dev & 617 & 1,795 & 672 & 3,084 \\
        Test & 474 & 1,423 & 598 & 2,495 \\
        \hline
    \end{tabular}
    
    \label{tab:dataset_statistics}
\end{table}
After deduplication, our dataset has \textbf{15,514} claims. The dataset is unbalanced, favoring refuted or false claims with a distribution of `True', `False', and `Conflicting' claims at 18.79\%, 57.93\%, and 23.27\%, respectively, reflecting the tendency of fact-checkers to focus on false information \cite{schlichtkrull2023averitec}. A detailed distribution of the dataset is shown in Table \ref{tab:dataset_statistics}. 

To illustrate the comprehensive nature of our dataset, we analyze on the origins of the claims it encompasses. Table \ref{tab:claim_sources} presents a summary of the ten most frequently encountered fact-checking websites within our dataset. It is noteworthy that Politifact constitutes a substantial fraction of the claims. Additionally, our analysis reveals a significant representation of claims sourced from a variety of fact-checking platforms, extending beyond the realm of political fact-checks. Furthermore, as illustrated in Table \ref{tab:georgraphical_sources}, our dataset encompasses claims originating from a wide range of geographical locales, including, North America, Europe, Asia, and Africa, underscoring the geographical diversity of our corpus.

\begin{figure}
    \centering
    \includegraphics[width=0.4\textwidth]{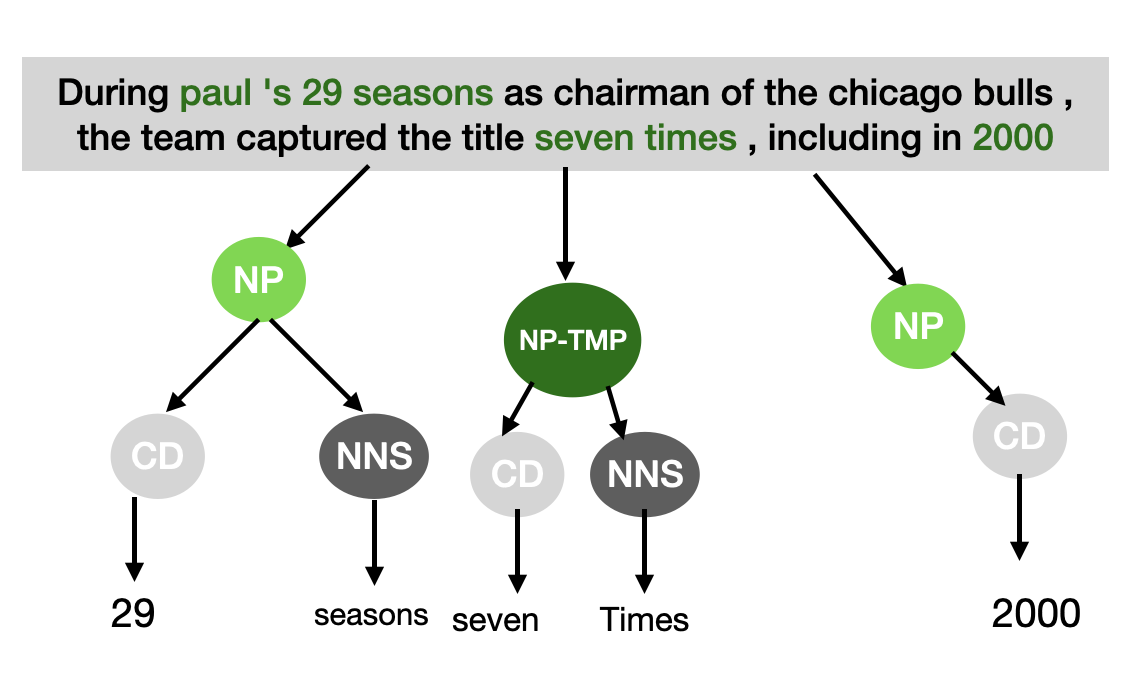}
    \caption{Example of identification of quantitative segments from the claim. NNS is noun plural form, NP-TMP is temporal noun phrase}
    \label{fig:quantitative_segments}
\end{figure}
\subsection{Identifying Quantitative Segments}
\label{section:quantitative}
We identify quantitative segments in the claim sentence for extracting numerical claims, as defined in \cite{ravichander-etal-2019-equate}, which include numbers, units, and optionally approximators (e.g.,``roughly'') or trend indicators (e.g., ``increases'').

Specifically, we first obtain the claim's constituency parse, identifying nodes with the cardinal number POS tag ``CD''. To avoid false positives (for example: ``The one and only''), we then parse these nodes' ancestors and extract noun phrases from their least common ancestors. Using these noun phrases as root nodes, we perform a prefix traversal of their subtrees. Figure \ref{fig:quantitative_segments} shows an example of the extracted quantitative segments. We then refine the claim set by filtering for those with at least one quantitative segment.


This approach is limited, as it may include claims with non-quantitative terms like ``Covid-19''. To remedy this, we require more than one quantitative segment, excluding any nouns like ``Covid-19'' mentions, to qualify as a numerical claim. Claims not meeting this criterion are excluded. Our self-assessment of 1000 sample claims from the dataset indicates a 95\% accuracy of our heuristic.


\subsection{Claim Decomposition}
We posit that the precise nature of the task of verifying numerical information requires the retrieval of relevant evidence containing the required quantitative information. Claim decomposition \cite{programfc, claim_decompose, claimdecomp, QABriefs} is effective in extracting important evidence containing background or implied information to verify the claim. Hence, we evaluate several claim decomposition approaches as described in Section \ref{section:decomposition} and discuss their impact on downstream veracity prediction in Section \ref{section:results_decomp}. We also release the relevant questions resulting from the decomposition of the claim for train, validation, and test sets to facilitate further research on their significance for retrieving relevant evidence.
\subsection{Collecting Evidence}
\label{section:evidence}

We collect evidence for claim verification by retrieving search results from public engines, following prior works \cite{multifc,schlichtkrull2023averitec}. Solely using the claim as a query, as in prior approaches \cite{multifc,popat2018declare,Botnevik:2020:SIGIR}, yields a restricted set of documents, often biased towards claims that have already been verified. To enhance the diversity of the evidence, we include both the original claim and questions generated from the decomposition methods, described in Section \ref{section:decomposition}. We get inspiration from recent advances in claim decomposition methods for fact-checking that have shown promising results~\cite{QABriefs,programfc,schlichtkrull2023averitec}.
To ensure the evidence collection is free from duplicates, we merge the search results for the original claim and all the generated questions using a top-k document pooling method commonly used in information retrieval (IR) \cite{pooled_1,pooled_2}. 

We first submit original claims to Google through \url{scaleserp.com} API, collecting the top 10 results per claim. We strictly filter out any results from over 150 fact-checking domains to prevent any leakage from their justification, avoiding models to learn shortcuts in verification. We improve evidence diversity by employing Large Language Model (LLM)-based claim decomposition methods such as \programfc~\cite{programfc} and \claimdecomp~\cite{claimdecomp}, generating a range of questions for query use. For each question, we compile the top 10 search results.

After filtering out duplicates and irrelevant documents, we amass an extensive evidence collection of 423,320 snippets. We have evidence from a diverse set of domains including Wikipedia, government websites, etc. An overview of top evidence domains is shown in Table \ref{tab:evidence_domains}. We ensure our collection does not have any snippets from manual or automated fact-checkers and related websites or social media handles. Also, government websites are one of the frequently occurring domains in our evidence collection, as our claims comprise diverse political and international events. Not surprisingly, many evidence snippets are from \texttt{statisa.com} and \texttt{investopedia.com} since our dataset is focused on numerical claims. On the other hand, we have 2,889 snippets from \texttt{youtube.com} which is surprising. After closer inspection, these snippets contain transcripts from news videos and textual descriptions which that relevant to the claim.

\subsection{Qualitative Analysis of the Dataset}
\label{section:manual_evaluation}

To ensure that the dataset we collect is of good quality, we perform automated and manual evaluations. Specifically, in manual evaluation, we evaluate the usefulness and comprehensiveness of the generated questions by various claim decomposition methods and the usefulness of the evidence they yield towards verifying the claim. We detail the guidelines used in the annotation process in this section.

We rate the questions generated on 2 aspects: completeness and usefulness (only based on the claim given and top search results). The two annotators are trained computer scientists who are closely associated with the task of automated fact checking and are familiar with the domain. The following guidelines were provided to the annotators. 

\textbf{Completeness}: A list of questions is said to be complete if the questions cover all aspects of the claim. The ratings were carried out according to the Likert scale of 1-5.

\textbf{Usefulness}: Rating the usefulness of questions determines if the questions would help verify the claim. The annotators were asked to rate the usefulness of questions resulting from the decomposition of claims. They were instructed to consider implicit aspects of the claim when rating this and were asked to rate according to the Likert scale 1-5.

Some questions might be relevant, but may just retrieve background knowledge and may not be relevant to the core aspect being fact checked. To gauge this, the annotators were also asked to look at evidence when rating the usefulness of questions. 

While evaluating the usefulness, the annotators were asked not to make assumptions about verification method. They were instructed to check if the questions covered all aspects of the claim and could retrieve relevant evidence. They were also asked to check the coverage of the implied meaning of the claim rather than just a surface level analysis of the claim. For instance, the implied aspect in the claim ``The ICE spends 4 times the amount to detain a person for a year than on a student in public school.
" is that ICE detains people for more than a year, which is not the case in reality. To verify this core aspect of the claim, the questions must be useful to retrieve related evidence that cites the time usually people are detained by the ICE.

\textbf{Evidence Usefulness}: The annotators were requested to rate the individual piece of evidences for each question. The usefulness depends on the information contained in the evidence. The information should be sufficient to support the whole or parts of the claim and should be rated on the Likert scale of 1-5 based on the degree of information. They were asked to rate the usefulness of all evidence by aggregating the usefulness of evidence tied to individual questions. The annotators were asked to rate individual evidence based on relatedness to the claim and their utility in fact checking.

\section{Experimental Setup}
\label{section:experiments}
To evaluate the \numclaims{} dataset, we introduce a baseline fact-checking pipeline. We fix the retriever model (BM25 + re-ranking) for all experiments. After extensive experiments, we choose to fine-tune the Roberta-Large-MNLI\footnote{\url{https://huggingface.co/roberta-large-mnli}} model, pre-fine-tuned on the MNLI corpus, for the NLI task. In Section \ref{subsec:stance_ablation} we further explore various NLI models' effectiveness on numerical claims.
\subsection{Claim Decomposition}
\label{section:decomposition}

We evaluate the following claim decomposition approaches. 

\mpara{\claimdecomp{}} \cite{claimdecomp}: The authors of this work provide annotated yes/no sub-questions for the original claims. We use \gptt{} on training samples from the \claimdecomp{} dataset to create yes/no sub-questions for our \numclaims{} dataset through in-context learning, setting temperature to 0.3, frequency to 0.6 and presence penalties to 0.8.

\mpara{\programfc} \cite{programfc}: We implement the approach proposed in this work to decompose claims and generate programs to aid in verification. The programs are step by step instructions resulting from decomposition of original claim. We employ gpt-3.5-turbo for decomposition. We use same hyperparameters as in the original paper.

\mpara{Original Claim}: Here we do not employ any claim decomposition, but rather use the original claim to retrieve evidence for arriving at the final verdict using the NLI models.




\begin{table}[hbt!]
 \caption{A broad overview of different categories of claims in \numclaims{}}
  \scalebox{0.9}
 {\begin{tabular}{lp{5.5cm}l}
\hline
\textbf{Category} & \textbf{Examples} & \textbf
{\#of claims}   \\ \hline

\midrule
Statistical & We've got 7.2\% unemployment (in Ohio), but when you include the folks who have stopped looking for work, it's actually over 10\%. & 7302 (47.07\%)  \\

 \midrule
 Temporal & The 1974 comedy young frankenstein directly inspired the title for rock band aerosmiths song walk this way & 4193 (27.03\%) \\
\midrule
 Interval & In Austin, Texas, the average homeowner is paying about \$1,300 to \$1,400 just for recapture, meaning funds spent in non-Austin school districts& 2357 (15.19\%)\\
\midrule
 Comparison & A vaccine safety body has recorded 20 times more COVID jab adverse reactions than the government’s Therapeutic Goods Administration. &1645 (10.60\%)  \\
 \\
\bottomrule
\end{tabular}}

\label{tab:categories}
\end{table}

\subsection{Evidence retrieval and Veracity Prediction}

\label{section:fact_verification_pipeline}
Once we have decomposed claims, we use evidence retrieval and re-ranking. Then we employ a classifier fine-tuned on \numclaims{} for the NLI task to verify the claim.
The different settings in which we evaluate the approaches are:

\mpara{Unified Evidence}: Our experiments utilize the evidence snippets collection detailed in Section \ref{section:evidence}. For each question/claim, we retrieve the top-100 documents using BM25 and re-rank them with \textit{paraphrase-MiniLM-L6-v2} from sentence-transformers library \cite{reimers-2019-sentence-bert}, selecting the top 3 snippets for the NLI task. We experiment with different values of $k \in [1,3,5,7]$ for top-k snippets. We observe that k=3 at claim level works best for our setup. Hence, for the ``Original Claim'' baseline, we use the top 3 evidences using the claim, and for other methods, we use the top-1 evidence per question, ensuring three evidences per claim for fair comparison. We experiment with different encoder models in huggingface such as Deberta-base, Deberta-large, Roberta family of models and observe that the 6-layer MiniLM model \cite{MiniLm} (\textit{paraphrase-MiniLM-L6-v2}) provides the best evidence ranking for downstream fact-verification.


After retrieving evidence, we fine-tune a three-class classifier for veracity prediction. Training and validation sets are formed using the retrieved evidence (described above), and the classifier is fine-tuned by concatenating the claim, questions, and evidence with separators, targeting the claim veracity label.

\mpara{Gold Evidence}: Here, we directly employ the justification paragraphs collected from fact-checking sites as evidence to check the upper bound for performance.

\mpara{Hyperparameters}: All classifiers are fine-tuned till ``EarlyStopping'' with patience of 2 and batch size of 16. AdamW optimizer is employed with a learning rate of $2e-5$ and $\epsilon$ of $1e-8$ and linear schedule with warm up. We use transformers library \cite{wolf2020huggingfaces} for our experiments.

\subsection{Category Assignment}
\label{section:taxonomy}
After curating the numerical claims, we categorize the numerical claims to one of these categories using a weak supervision approach. We identify four categories: temporal (time-related), statistical (quantity or statistic-based), interval (range-specific), and comparison (requiring quantity comparison) claims. The examples and distribution of these categories are shown in Table \ref{tab:categories}. We perform this categorization to aid in a fine-grained analysis of performance across different dimensions of numerical claims, as described in Section \ref{section:categories}. The categorization would help understand the performance of existing models on different types of numerical claims and gauge the scope for improvement.

We first manually annotate 50 claims, then used a few samples as in-context examples for the gpt-3.5-turbo model to label hundreds more. After initial labeling, we fine-tuned Setfit \cite{tunstall2022efficient}, a classifier ideal for small sample sizes on the annotated samples. Then we employed the classifier for further annotation of the entire dataset to the defined categories. Two annotators manually reviewed 250 random claims,
and 199 claims were found to be  correctly categorized. These annotations helped refine the classifier and category assignment.

\subsection{Veracity Prediction Model Ablations}
\label{subsec:stance_ablation}

\textbf{Prompting based Generative NLI models}:
We assess the stance detection using large generative models like \flant{} (3B params) and \gptt{}, providing them with training samples, ground truth labels, and retrieved evidence as in-context examples for claim verification. The models are also prompted to produce claim veracity and justification jointly to ensure faithfulness, with a temperature setting of 0.3 to reduce randomness in outputs. 
\begin{table*}[hbt!]

    \centering
     \caption{Results of different models on \numclaims{} (categorical and full) with Roberta-Large-MNLI as the NLI model. M-F1 : Macro-F1, W-F1 : Weighted-F1 and C-F1 refers to F1 score for Conflicting class.$\dagger$ indicates statistical significance at 0.01 level over baseline using t-test}
    \begin{tabular}{lccccccccccccll}
    \toprule
     \textbf{Method}&\multicolumn{2}{c}{Statistical} & \multicolumn{2}{c}{Temporal} & \multicolumn{2}{c}{Interval} & \multicolumn{2}{c}{Comparison}& \multicolumn{3}{c}{Per-class F1} &\multicolumn{2}{c}{\numclaims{}} \\
   &\multicolumn{1}{c|}{M-F1} & \multicolumn{1}{c}{W-F1}&\multicolumn{1}{c|}{M-F1} & \multicolumn{1}{c}{W-F1}&\multicolumn{1}{c|}{M-F1} & \multicolumn{1}{c}{W-F1}&\multicolumn{1}{c|}{M-F1} & \multicolumn{1}{c}{W-F1} & \multicolumn{1}{c|}{T-F1} & \multicolumn{1}{c|}{F-F1} & \multicolumn{1}{c}{C-F1} & \multicolumn{1}{c|}{M-F1} & \multicolumn{1}{l}{W-F1} \\
     \midrule
    
    \textbf{Unified Evidence Corpus} \\
    Original Claim (baseline) & 49.55 & 52.48 & 60.29& 74.29& 48.84 & 57.93 & 40.72 & 39.66& 51.59& 70.60&35.27 & 52.48 & 58.52 \\
    \programfc{} & 52.24 & 57.83 & 56.75 & 75.46& 47.09 & 61.88& 49.02& 48.07& 47.42& 79.46 & 33.40 & 53.43 & 62.34
    \\

    \claimdecomp{} & 53.34 & 58.79 & 61.46 & 78.02 & 56.02 & 66.97 & 53.59 & 53.44 & 51.82 & 79.82 & 39.72 & \textbf{57.12} $\dagger$ & \textbf{64.89}$\dagger$
    \\

     \midrule
    \textbf{Fine-tuned} \\
    \textbf{w/ Gold Evidence}\\
    \numclaims{} only & 60.87 & 65.44 & 66.63 & 81.11 & 58.35 & 69.56 & 60.74 & 60.36 & 56.86 & 82.92 & 48.79 & \textbf{62.85} & \textbf{69.79}
    \\
    
    \numclaims{} + non-num & 56.76 &  61.98 & 64.04& 80.35 & 56.56 & 67.13 & 52.03 &50.59  & 59.87 & 83.13 & 33.78 & 58.66 & 66.73
    \\ 
    \midrule
    Naive (Majority class) &22.46 & 34.25 & 28.35 & 62.95 & 25.86 & 49.19 & 16.51& 16.32& 0.00&  72.64 & 0.00  &24.21 & 41.42
    \\
    \bottomrule
    \end{tabular}
   
    \label{tab:main_result}
\end{table*}
The few-shot prompt employed for veracity prediction through \gptt{} is shown in Table \ref{prompt_veracity}. We dynamically select few shot examples for every test example. The examples shown are for a single instance and are not indicative of the examples used for inference over all test examples. We also show the results for zero-shot veracity prediction using generative models. 
\subsubsection{Fine-tuned models} We fine-tune T5-small (60 M params), \bart{} and Roberta-large (355 M params) to study the impact of scaling on verifying numerical claims.
We also employ models pre-trained on number understanding tasks such as \finqaroberta{} \cite{zhang2022elastic}, \numt{} \cite{yang2021nt5}. We fine-tuned these models on our dataset for the NLI task to test the hypothesis if models trained to understand numbers better aid in verifying numerical claims.
All models are fine-tuned with hyperparameters described in Section \ref{section:fact_verification_pipeline}.

\section{Results}
\subsection{Automated Analysis of Quality of decomposition}
Examining questions generated by \claimdecomp{} and \programfc{}, we prioritize their relevance to the original claim and diversity in covering different claim aspects. 
BERTScore \cite{zhang2020bertscore} is employed to assess relevance, i.e., measuring how well the questions align with the claims. For diversity, which ensures non-redundancy and coverage of various claim facets, we utilize the sum of (1-BLEU) and Word Position Deviation \cite{wpd}. The results are shown in Table \ref{tab:automated_question_quality}. Our findings indicate that \claimdecomp{} excels in generating questions that are not only more relevant but also more diverse compared to \programfc{}, addressing multiple facets of the claim.
We also perform manual analysis by sampling 20 claims sampled from test set along with decomposed questions and retrieved evidence for \programfc{} and \claimdecomp{} approaches.

\subsection{Manual Evaluation Results}
We perform a manual evaluation of questions from decomposed claims and retrieved evidence in \numclaims{} as described in Section \ref{section:manual_evaluation}. We ask two computer scientists familiar with the field to annotate them on measures of completeness (if questions cover all aspects of the claim), question usefulness and evidence usefulness, where usefulness is measured by information they provide to verify the claim. The results are shown in Table \ref{tab:question_quality}.
We also report the Cohen's kappa scores, indicating the level of agreement among the annotators in the table. We observe that questions generated by \claimdecomp{} are better than \programfc{} as measured by their usefulness and quality of evidence retrieved. We also observe that the decompositions yielded by both the approaches cover all aspects of the claim, as observed by the completeness score. We also release the questions from claim decomposition for train, validation, and test sets of \numclaims{}.
\subsection{Hardness of Numerical Claim Verification}
To address \textbf{RQ1}, we experiment with various claim verification approaches on the \numclaims{} dataset, considering both unified evidence and gold evidence. The performance of different approaches is presented in Table~\ref{tab:main_result}. \textit{Fine-tune using Gold Evidence:} shows performance of NLI models fine-tuned on \numclaims{} (numerical only) and \numclaims{}+non-num (numerical and non-numerical claims) using gold evidence snippet. Both numerical claims and non-numerical claims are from the same fact-checkers. 

It is evident that \numclaims{} poses a considerable challenge for fact-checking numerical claims, with the best approach achieving a weighted-F1 of $64.89$ for unified evidence and $69.79$ for gold evidence. 
The difficulty is further underscored by the performance of the naive baseline, which simply predicts the majority class. 
A similar trend is observed at the categorical level. 
Except for the temporal category, where it outperforms other categories, the improvements from the baseline is relatively modest. 

Additionally, training specifically on \numclaims{}'s numerical claim distribution improves performance by 7.14\% in macro F1 compared to a mixed distribution of numerical and non-numerical claim set. These results underscore the complexity of verifying numerical claims.
\begin{table*}[hbt!]
    \centering
        \caption{Ablation results employing different NLI models for \claimdecomp{} on  \numclaims{}. The best results are in \textbf{bold}.$\dagger$ indicates statistical significance at 0.01 level over Roberta-large using t-test}
    \begin{tabular}{lcccccccccccccc}
    \toprule
     \textbf{Method}&\multicolumn{2}{c}{Statistical} & \multicolumn{2}{c}{Temporal} & \multicolumn{2}{c}{Interval} & \multicolumn{2}{c}{Comparison}& \multicolumn{3}{c}{Per-class F1} &\multicolumn{2}{c}{\numclaims{}} \\
   &\multicolumn{1}{c|}{M-F1} & \multicolumn{1}{c}{W-F1}&\multicolumn{1}{c|}{M-F1} & \multicolumn{1}{c}{W-F1}&\multicolumn{1}{c|}{M-F1} & \multicolumn{1}{c}{W-F1}&\multicolumn{1}{c|}{M-F1} & \multicolumn{1}{c}{W-F1} & \multicolumn{1}{c|}{T-F1} & \multicolumn{1}{c|}{F-F1} & \multicolumn{1}{c}{C-F1} & \multicolumn{1}{c|}{M-F1} & \multicolumn{1}{c}{W-F1} \\
     \midrule

    \midrule
    
    \textbf{Unified Evidence Corpus} & & \\

         \bart{} & 52.89 & 58.43 & \textbf{62.01}& \textbf{78.07} & 54.52 & 65.85 & 53.63 & 53.49&51.23& 79.56 & 39.37 & 56.71 &64.54  \\

         Roberta-large & 53.56 & 58.24 & 59.31 & 75.67& 51.64 & 62.38 & 43.91 & 42.39& 50.58 & 77.23 &35.50&54.43 & 62.16 \\
         T5-small & 43.52 & 51.37 & 32.08 & 64.65 & 43.64 & 60.38& 48.41 & 48.88 &  19.65 & 77.22 & 38.02 &  44.96 & 56.89 \\
         \numt{} & 50.36 & 56.58 & 41.35 & 68.59 & 47.96 & 61.35 & 49.14 & 48.90 & 36.56 & 78.45 & 35.76 & 50.26 & 60.26  \\
         \finqaroberta{} & \textbf{56.97} & \textbf{61.36} & 60.29 & 75.55& \textbf{56.53} & \textbf{66.52} & 52.53 & 52.34 & 49.72 & 77.91& \textbf{47.33} & \textbf{58.32}$\dagger$ & \textbf{65.23}$\dagger$  \\

        FlanT5 (zero-shot) & 32.11 & 36.43 & 26.51 & 42.64& 32.36 &44.03 & 27.48 & 24.95 & 36.35 & 52.56& 3.15 & 30.68 & 37.64\\

          FlanT5 (few-shot) & 37.31 & 41.24 & 32.70 & 46.83& 37.61 &47.52 & 35.20 & 34.47 & 33.90 & 54.73& 20.92 & 36.52 & 42.67\\
    \gptt{} (zero-shot) & 34.51 & 33.78 & 28.04 & 29.12& 35.34 &36.98 & 40.31 & 40.45 & 37.81 & 32.57& 31.25 & 33.87 & 33.25\\

        GPT4 (few-shot) & 37.04 & 41.24 & 31.90& 46.29& 37.52 &47.88 & 37.16 & 39.41 & 14.38 & 52.82& 42.31 & 36.50 & 42.99\\
 \gptt{} (few-shot) & 48.26 & 51.93 & 42.90 & 57.67& 43.41 & 54.10 & 45.84 & 45.29& 44.41 & 64.26 & 32.35& 47.00 & 50.98 \\
 
     


     \midrule
\textbf{Gold Evidence} \\

\gptt{} (few-shot) &53.40 & 57.51 & 50.88 &69.15& 50.97 & 62.10 & 51.05 & 49.56  & 56.77 & 75.35 & 28.00  & 53.37 & 60.47\\ 
 \midrule


     \bottomrule
    \end{tabular}

    \label{appendix:ablations}
\end{table*}
\begin{table}[t!]
 \small
 \centering
\caption{Manual evaluation of decomposed questions. C: Completeness, QU : Question usefulness, EU : Evidence Usefulness. We use the Likert scale of 1-5 and report Cohen's kappa ($\kappa$) for inter-annotator agreement.}
\begin{tabular}{lccc}
\hline
\textbf{Method}&\textbf{C} ($\kappa$)& \textbf{QU} ($\kappa$) & \textbf{EU} ($\kappa$)\\ 
\hline

\midrule
\programfc{} & 4.6 $\pm$0.77 (0.65) & 3.4 $\pm$1.15 (0.53) & 2.9 $\pm$1.74 (0.66)  \\

\claimdecomp{} & 4.5 $\pm$0.86 (0.70) & \textbf{3.7 $\pm$0.92} (0.59)  & \textbf{3.2 $\pm$1.41} (0.69) \\

 
\bottomrule
\end{tabular}

\label{tab:question_quality}
\end{table}



 

\begin{table}[hbt!]
 \centering
 \caption{Automated evaluation of decomposed questions.}
\begin{tabular}{lcc}
\toprule
\textbf{Method}  & \textbf{Relevance} & \textbf{Diversity} \\ 
\midrule
\programfc{}  & 0.782 & 0.430  \\
\claimdecomp{}  & \textbf{0.831} & \textbf{0.490} \\

\bottomrule
\end{tabular}

\label{tab:automated_question_quality}
\end{table}
\subsection{Effect of Claim Decomposition on Claim Verification}
\label{section:results_decomp}
The \textbf{RQ2} is answered by Table \ref{tab:main_result} which indicates that claim decomposition enhances claim verification, particularly for the 'Conflicting' category, where \claimdecomp{} outperforms original claim-based verification significantly. 
In the 'unified evidence' setting, \claimdecomp{} sees gains of $8.84\%$ in macro-F1 and $10.9\%$ in weighted-F1. This improvement is attributed to more effective evidence retrieval for partially correct claims, as supported by categorical performance. Using original claims sometimes leads to incomplete or null evidence sets. Numerical claim verification requires multiple reasoning steps, as seen in Example 1 from Table \ref{tab:qualitative}. Claim decomposition creates a stepwise reasoning path by generating questions on various aspects of the claim, thereby providing necessary information for verification.


\subsection{Effect of Different NLI models}
To assess the impact of different NLI models, we utilize \claimdecomp{}, the top-performing claim decomposition method from Table \ref{tab:main_result}. Table \ref{appendix:ablations} addresses \textbf{RQ3}, showing that models trained on numerical understanding, like \numt{} and \finqaroberta{}, surpass those trained only on general language tasks. Specifically, \numt{} beats T5-small by $11.8\%$ in macro F1, and \finqaroberta{}, a number-focused Roberta-Large model, exceeds the standard Roberta-Large model by the same margin. The highest performance is achieved by \finqaroberta{}, which also outperforms Roberta-large-MNLI.

Finally, we address \textbf{RQ4} by studying the model scale's impact on claim verification reveals that larger models improve performance when fine-tuned, but not necessarily in few-shot or zero-shot settings. For example, \gptt{} under-performs in few-shot and zero-shot scenarios compared to smaller fine-tuned models ($355M$ or $60M$ parameters). This under-performance, observed in \flant{} and \gptt{}, is often due to hallucination, where models incorrectly interpret evidence or reach wrong conclusions about claim veracity despite parsing accurate information.
\vspace{-1em}
\begin{table*}
\centering
\caption{Qualitative analysis of results from different claim decomposition approaches}
\begin{tabular}{lp{15cm}}
\toprule
    \textbf{Method}     & \textbf{Decomposition and Verdict}  \\
\midrule
 \colorg \textbf{Claim}  & \colorg \textbf{Discretionary spending has increased over 20-some percent in two years if you don’t include the stimulus. If you put in the stimulus, it’s over 80 percent} \\
Original Claim & \textbf{[Verdict]: \textcolor{magenta}{True}}  \\

\claimdecomp{} & \textbf{[Decomposition]:} [Q1]:Has discretionary spending increased in the past two years?,[Q2]:Does the increase in discretionary spending exclude the stimulus? [Q3]: Is there evidence to support the claim that  \dots \textbf{[Verdict]: \textcolor{teal}{Conflicting}} \\

\programfc{} &    \textbf{[Decomposition]:}     fact\_1 = Verify(Discretionary spending has increased over 20-some\dots),
        fact\_2 = Verify(``If you don't include \dots, discretionary spending has increased\dots"),
        fact\_3 = Verify(``If you put in the stimulus, discretionary spending\dots"),
        \textbf{[Verdict]: \textcolor{magenta}{True}} \\

\midrule
 \colorg \textbf{Claim}  & \colorg \textbf{Under GOP plan, U.S. families making ~\$86k see avg tax increase of \$794.} \\

Original Claim & \textbf{[Verdict]: \textcolor{magenta}{Conflicting}}  \\
\claimdecomp{} & \textbf{[Decomposition]:} [Q1]:is the tax increase under the gop plan in the range of \$794 \dots making about \$86,000?,[Q2]:does the gop plan result in an average tax increase\dots \$86,000?[Q3]:is there evidence that\dots? 
\textbf{[Verdict]: \textcolor{teal}{False}}   \\

\programfc{} &    \textbf{[Decomposition]:}     fact\_1 = Verify(``Under GOP plan, U.S. families making ~\$86k\dots") 
        \textbf{[Verdict]: \textcolor{magenta}{Conflicting}} \\
\bottomrule
\end{tabular}

\label{tab:qualitative}
\end{table*}

\begin{table*}[h!!!]
\small
\caption{Example of In-context learning sample for \gptt{} few shot for claim verification}
\begin{tabular}{p{17cm}}
\toprule 
\textbf{Prompts for few-shot fact verification with \gptt{}. } Note , the ICL samples are selected dynamically for each test sample:
\\
\midrule


\textbf{System prompt}: Following given examples,  For the given claim and evidence
        fact-check the claim using the evidence
        , generate justification and output the label in the end.
        Classify the claim by predicting the Label: in the end
        as one of: 
        SUPPORTS, REFUTES or CONFLICTING.\\

\textbf{User Prompt}:


\textbf{[Claim]:}"A family of four can make up to \$88,000 a year and still get a subsidy for health insurance" under the new federal health care law. \\
\textbf{[Questions]:}is there a subsidy for health insurance under the new federal health care law?
  can a family of four with an income of \$88,000 a year qualify for a subsidy for health insurance?
  does the new federal health care law provide subsidies for families with an income of \$88,000 a year? \\
\textbf{[Evidences]:}by the way, before this law, before obamacare, \dots make for example, a family of four earning \$80,000 per year  ... \\
\textbf{Label:}SUPPORTS \\
\dots 
              Following given examples,  
        for the given claim,  given questions and evidence
        use information from them to
        fact-check the claim and also additionally paying attention
        to highlighted numerical spans in claim and evidence. Input [\dots]
                      
\\
\bottomrule
\end{tabular}

 \label{prompt_veracity}
\end{table*}

\subsection{Performance across different categories of numerical claims}
\label{section:categories}
We assessed our fact-checking pipeline's limitations by evaluating baselines in the four categories detailed in Section \ref{section:taxonomy}. Table \ref{tab:main_result} shows that methods like \claimdecomp{}, which use claim decomposition, outperform original claim-based verification in all categories. Specifically, for comparison based claims \claimdecomp{} sees gains of $34.7\%$ in weighted F1 and $31.6\%$ in macro F1 over original claim verification. This is particularly effective for comparison and interval claims, where decomposition aids in handling claims requiring quantity comparisons or reasoning over value ranges, resulting in better evidence retrieval.

In our analysis of different NLI models, fine-tuned models show better performance across all four categories with increased scale. Notably, models with a focus on number understanding, like \numt{} and \finqaroberta{}, outperform those trained only on language tasks. This is especially relevant for statistical claims that often require step-by-step lookup and numerical reasoning, where \finqaroberta{} achieves a weighted F-1 of \textbf{61.36}. Although decomposition approaches and number understanding NLI models enhance performance, explicit numerical reasoning is key for further improvements, a topic for future exploration. A detailed ablation study with macro and weighted F1 scores for all categories of numerical claims for different NLI models is shown in Table \ref{appendix:ablations}.

\subsection{Error Analysis}
\label{subsec:error_analysis}
We conduct an analysis of claims in the test set and their corresponding predictions, offering insights into the considered fact-checking pipeline. Examining results in Table \ref{tab:main_result} and Table \ref{appendix:ablations}, we note the challenge in verifying claims categorized as "conflicting." These claims pose difficulty as they are partially incorrect, requiring the retrieval of contrasting evidence for different aspects of the original claim. We also observe that NLI models with numerical understanding, coupled with claim decomposition, yield better performance. However, there is room for further improvement, as the highest score for this class is only \textbf{47.33}.

Among other categories, we observe comparison based numerical claims to be the hard as they are mostly compositional and require decomposition around quantities of the claim followed by reasoning over the different quantities. While claim decomposition helps advance the performance by a significant margin of 31.6\% (macro F-1) (Table \ref{tab:main_result}), there are few errors in the decomposition pipeline for approaches like \programfc{}. For instance, claim decomposition may result in \textbf{over-specification} where the claim is decomposed to minute granularity or \textbf{under-specification} where the claim is not decomposed sufficiently. An example of over-specification is shown in the first example for \programfc{} in Table \ref{tab:qualitative} where the claim is over decomposed leading to an erroneous prediction. The second example demonstrates a case of under-specification where the claim is not decomposed, leading to limited information and erroneous verdict.

\section{Conclusions}
\label{section:conclusion}
We introduce \numclaims{}, the largest real-world fact-checking dataset to date, featuring \textit{numerical data} from global fact-checking sites. 
Our baseline system for numerical fact-checking, informed by information retrieval and fact-checking best practices, reveals that claim decomposition, models pre fine-tuned using MNLI data, and models specialized in numerical understanding enhance performance for numerical claims. 
We show that \numclaims{} is a challenging dataset for a variety of existing fine-tuned and prompting-based baselines.
\appendix

\section{Acknowledgements}
This work is in part funded by the Research Council of Norway project EXPLAIN (grant number 337133). We also thank Factiverse AI for providing the claim dataset.




\newpage

\bibliographystyle{ACM-Reference-Format}
\bibliography{sample-base}

\newpage
\appendix
\end{document}